\documentclass[11pt, letterpaper]{cmu}

\usepackage[utf8]{inputenc}
\usepackage[english]{babel}
\usepackage{amsthm,amsmath,amsfonts,amssymb}
\usepackage{cmap}
\usepackage[T1]{fontenc}
\usepackage{url}
\usepackage[colorlinks=true,allcolors=blue]{hyperref}
\usepackage[dvipsnames]{xcolor}
\usepackage{textcomp}
\usepackage[overload]{textcase}
\usepackage{graphicx}
\usepackage{colortbl}
\usepackage{booktabs}
\usepackage{changepage}
\usepackage{enumitem}
\usepackage{tabularx}
\usepackage{longtable}
\usepackage{fancyhdr}
\usepackage[explicit]{titlesec}
\usepackage{caption}
\usepackage{XCharter}
\usepackage[scaled=1.1]{zlmtt}
\usepackage[
    left=2.2cm,
    right=2.2cm,
    top=3cm,
    bottom=3cm,
    headheight=40pt,
    headsep=20pt,
    letterpaper
]{geometry}
\usepackage[comma,numbers,sort,compress]{natbib}
\usepackage{microtype}
\usepackage{float}
\usepackage{mathtools}
\usepackage{mathrsfs}
\usepackage{multirow}
\usepackage{subcaption}
\usepackage{algorithm}
\usepackage{algorithmic}
\usepackage{array}
\usepackage{lineno}
\usepackage{etoolbox}
\usepackage{arydshln}

\AtBeginDocument{\expandafter\let\csname endtabular*\endcsname\endtabular}

\setlist[itemize]{noitemsep}
\setlist[enumerate]{noitemsep}
\definecolor{BerkeleyBlue}{HTML}{C41230}

\fancypagestyle{firststyle}{
    \fancyhead[L,C,R]{}
    \fancyfoot[L,C,R]{}
    \fancyfoot[L]{%
        \footerfont
        \textbf{Corresponding author:} \the\correspondingauthor
    }
    \renewcommand{\headrulewidth}{1pt}
    \renewcommand{\footrulewidth}{1pt}
}
\renewcommand{\headrulewidth}{1pt}
\renewcommand{\footrulewidth}{0pt}
\usepackage{wrapfig}

\makeatletter
\renewcommand{\maketitle}{
    \check@mathfonts %
    \begingroup
    \setlength{\parindent}{0pt}
    \begin{adjustwidth}{0pt}{24pt}
        \begin{center}
            \vskip5pt
            {\raggedright\titlefont\@title\par}
            \vskip11pt
            {\raggedright\@author\par}
            \vskip-5pt
            {\color{black}\makebox[\linewidth][l]{\rule{\textwidth}{1pt}}}
            \vskip10pt
        \end{center}
    \end{adjustwidth}
    \endgroup
    \thispagestyle{firststyle}
}
\makeatother

\titleformat{\section}
    {\large\bfseries\headingfont}
    {\thesection.}
    {0.5em}
    {#1}
\titleformat{name=\section,numberless}
    {\large\bfseries\headingfont}
    {}
    {0em}
    {#1}
\titleformat{\subsection}
    {\bfseries}
    {\thesubsection.}
    {0.5em}
    {#1}
\titleformat{\subsubsection}
    {\bfseries\itshape}
    {\thesubsubsection.}
    {0.5em}
    {#1}
\titlespacing*{\section}{0pc}{3ex plus4pt minus3pt}{5pt}
\titlespacing*{\subsection}{0pc}{2.5ex plus3pt minus2pt}{2pt}
\titlespacing*{\subsubsection}{0pc}{2ex plus2.5pt minus1.5pt}{2pt}

\DeclareCaptionLabelSeparator{pipe}{:~}
\hypersetup{
    colorlinks = true,
    citecolor = {magenta},
    linkcolor = {blue},
    urlcolor = {blue},
    pdftitle = {Training Generalist Value Functions for Long-Horizon Robotic Manipulation},
    pdfauthor = {Saksham Singh, Zheyuan Hu, Max Sobol Mark, Jeffrey Yu, Zackory Erickson, and Aviral Kumar},
}

\newcommand{\methodname}{\textsc{SeeQ}}

\newcommand{\msm}[1]{\textcolor{teal}{MSM: TODO}}

\usepackage{amsmath,amsfonts,bm}

\def\eqref#1{Eq.~\ref{#1}}

\def\1{\bm{1}}

\DeclareMathAlphabet{\mathsfit}{\encodingdefault}{\sfdefault}{m}{sl}
\SetMathAlphabet{\mathsfit}{bold}{\encodingdefault}{\sfdefault}{bx}{n}

\newcommand{\bs}{\mathbf{s}}
\newcommand{\ba}{\mathbf{a}}

\title{\methodname{}: Training Generalist Value Functions for Long-Horizon Robotic Manipulation}

\author[1]{Saksham Singh}
\author[1]{Zheyuan Hu}
\author[1]{Max Sobol Mark}
\author[1]{Jeffrey Yu}
\author[1]{Zackory Erickson}
\author[1]{Aviral Kumar}
\affil[1]{Carnegie Mellon University}

\correspondingauthor{\textit{saksham3@andrew.cmu.edu}. \textbf{Project website} with videos: \url{https://saksham002.github.io/seeq/}.\\ \textbf{Pretrained \methodname{} checkpoint}: \url{https://huggingface.co/CMU-AIRe/SeeQ-3B/}.}

\begin{document}
\maketitle

\vspace{-0.9cm}
\begin{center}
\begin{minipage}{0.99\textwidth}
  \centering
  \includegraphics[width=0.90\textwidth]{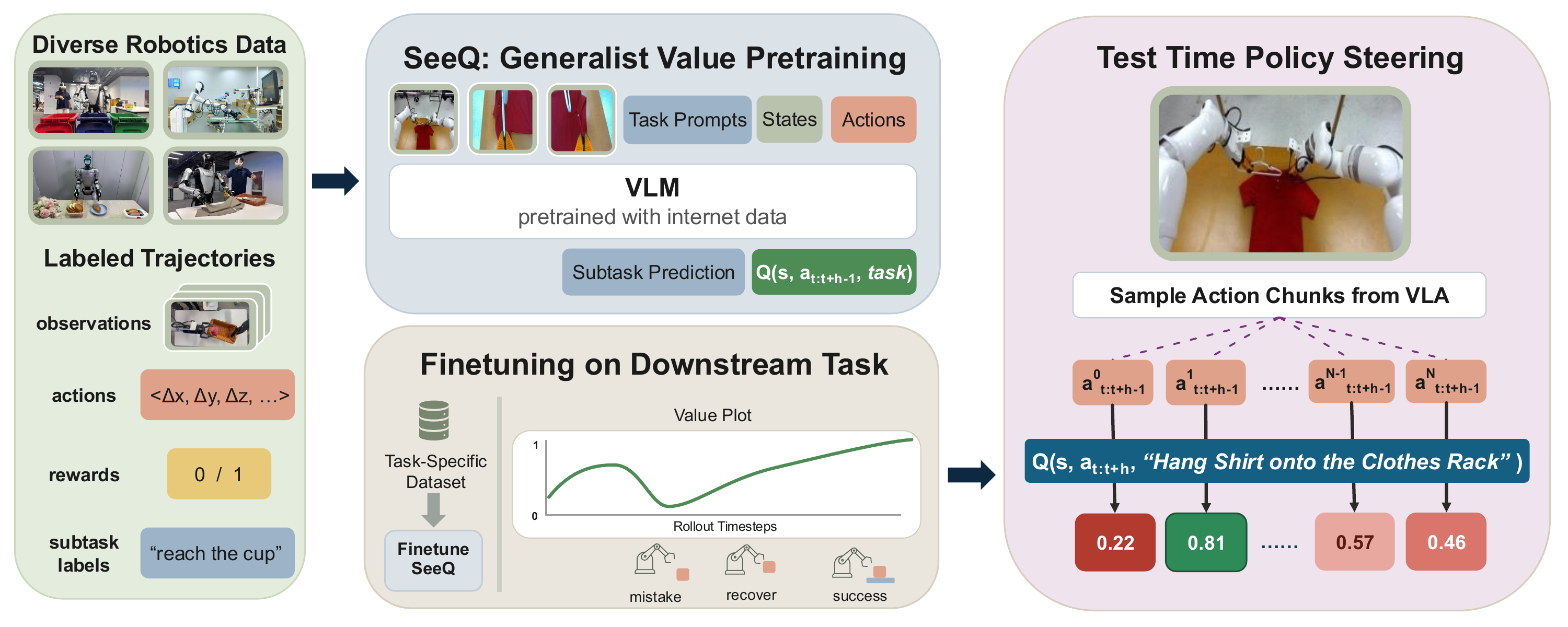}
  \vspace{-0.15cm}
  \captionsetup{hypcap=false}\captionof{figure}{\footnotesize{\textbf{\methodname{} overview.}
 We train a generalist Q-value function on diverse robot data using a pretrained vision-language model (VLM) backbone. The value function is then finetuned on a downstream task and used to steer a base policy at test time. To address the challenges of long-horizon value learning, we use temporal-difference (TD) learning to fit the Q-value of the \emph{currently active} subtask rather than that of the entire task. To eliminate the need for humans to specify this subtask at deployment, we modify the Q-function architecture to first predict the active subtask in natural language and then predict its value, encouraging internal representations better aligned with subtask-level value estimation.}}
  \label{fig:teaser}
\end{minipage}
\end{center}
\vspace{-0.2cm}

{\absfont
\noindent\textbf{Abstract:} Despite rapid progress, generalist robot policies remain brittle on complex, long-horizon tasks that comprise multiple stages or require repeated attempts and deliberation on the same underlying stage before success. Q-value functions can improve these policies by ranking candidate actions or guiding policy improvement, but learning from sparse task-level rewards entails long credit-assignment horizons, difficult Bellman backups, and broad data-coverage requirements. We introduce \methodname{} (\textbf{S}ubtask-\textbf{e}licit\textbf{e}d \textbf{Q}-functions), which instead learns Q-values for the currently active subtask. This shortens the value-prediction horizon and enables effective learning with temporal-difference (TD) objectives. During training, subtask-level annotations present in offline robot data provide the decomposition and enable learning from broad, potentially suboptimal robot datasets. To eliminate the need for human annotations or modular subtask prediction systems at test time, our Q-function architecture is trained to autoregressively predict the active subtask in natural language before estimating its value. We instantiate \methodname{} using a base vision-language backbone, pretrain it on diverse open-source robot manipulation data, and finetune it on downstream tasks. Across {four} real-world manipulation tasks on two bimanual robot platforms, the \methodname{} value function substantially improves best-of-$N$ policy steering.
}

\vspace{-0.3cm}
\section{Introduction}
\vspace{-0.2cm}

Generalist robot policies provide a promising foundation for robot control: they acquire useful perceptual and motor priors, follow language instructions, and transfer across objects, scenes, and embodiments~\citep{intelligence2025pi05visionlanguageactionmodelopenworld}. 
Yet robust autonomous deployment remains challenging, particularly for tasks that unfold over long horizons or demand high precision. Some long-horizon tasks may involve multiple stages, while others might simply be shorter precision-heavy tasks that require deliberating among similar actions and retrying after imperfect attempts. Imitation-trained policies struggle in both settings: errors compound over time, recovery behaviors are underrepresented in expert datasets, and behaviors of varying quality are modeled indiscriminately. Value functions offer a natural remedy by distinguishing actions that make progress from those that lead to failure, thereby steering a base policy toward more successful behavior~\citep{nakamoto2025steering}.

Recent work has shown that learned value functions can substantially improve imitation learning policies in the real world~\citep{mark2025policy,wang2026learningdeployingfleetscalereinforcement,intelligence2025pi06vlalearnsexperience}. Ideally, rather than learning a separate value function from scratch for each task, we would train a single \emph{generalist} Q-function that can improve policies across many tasks while benefiting from large-scale pretraining and diverse robot data. However, effective value-learning methods such as temporal-difference (TD) learning become unreliable over long horizons~\citep{park2026horizon}, precisely the regime in which complex tasks present the greatest challenges. Some approaches avoid this issue by discarding TD learning and directly regressing to Monte Carlo returns. Without TD learning, however, the value function cannot improve the policy far beyond the data collection policy, which limits the extent to which value learning can be helpful. This raises the central question of our work: \emph{can we build a generalist value learning approach that utilizes TD learning to learn from diverse data, sidesteps training over long-horizons, but can still produce value functions that are useful for long-horizon tasks?}

Our approach exploits a common structure in manipulation tasks: although a task specifies a distant outcome, the robot’s behavior at any moment is typically directed toward an immediate objective. Multi-stage tasks progress through intermediate milestones, such as grasping an object before placing it, while precision- or deliberation-heavy tasks may require repeated attempts to accomplish a given stage, but each attempt may follow a particular short-horizon objective. In both cases, evaluating progress toward the active subtask requires a shorter prediction horizon than evaluating success on the full task. Building on this structure, we train a generalist Q-function on a vision-language model (VLM) backbone and redefine its prediction target. Rather than modeling sparse success over the full task, the Q-function estimates the expected return for the currently active subtask. This yields \methodname{} (\textbf{S}ubtask-\textbf{e}licit\textbf{e}d \textbf{Q}-functions), which recasts long-horizon value learning as short-horizon, subtask-level value learning. To eliminate the need for subtask annotations at test time, we modify the Q-function to first predict the active subtask in natural language and then estimate its value. In addition, this objective improves robustness near subtask transitions and aligns value estimation with the pretrained VLM’s text-generation capabilities.

We instantiate \methodname{} using a pretrained 3B PaliGemma vision-language backbone~\citep{beyer2024paligemmaversatile3bvlm} and train it on diverse robot manipulation datasets with subtask annotations, including RoboCOIN~\citep{wu2026robocoinopensourcedbimanualrobotic}. We evaluate whether the resulting Q-function can infer subtasks for held-out instructions, adapt to new embodiments with limited finetuning, and improve fixed, generalist imitation-learned policies. Empirically, \methodname{} decomposes out-of-distribution tasks into meaningful subtask spans and predicts useful values for them, with the subtask-prediction loss proving critical to performance. When used for best-of-$N$ action selection, \methodname{} improves generalist robot policies~\citep{intelligence2025pi05visionlanguageactionmodelopenworld} on \textbf{four} real-world bimanual manipulation tasks across \textbf{two} robot platforms, including precision-heavy tasks with deformable objects requiring recovery and planning-heavy tasks comprising many stages. These results suggest that subtask-level values provide a more effective learning target than sparse, long-horizon success when training generalist Q-functions.

\vspace{-0.3cm}
\section{Preliminaries, Definitions, and Notation}
\label{sec:prelims}
\vspace{-0.2cm}
We formulate our problem in a language-conditioned robotic manipulation setting with access to a base generalist policy, $\pi_\text{base}$, that we wish to improve. In our experiments, we often finetune an off-the-shelf policy on in-domain teleoperation data. Given a natural language task instruction $l$ and a state $\bs_t$, specified by visual observations (e.g., over-the-shoulder and wrist cameras) and proprioceptive robot state, the policy produces an action chunk~\citep{zhao2023learningfinegrainedbimanualmanipulation}, i.e.,
$\ba_{t:t+H-1} \sim \pi_\text{base}(\cdot \mid \bs_t, l)$.
The environment then executes this chunk, or a short prefix of it, before replanning from a new state.

To improve this policy, typically we learn a Q-function that estimates the expected reward-to-go of an action chunk $\ba_{t:t+H-1}$ from state $\bs_t$, conditioned on the task instruction $l$. The instruction specifies a sparse reward function $r_l$, where $r_l(\bs_t)=1$ if task $l$ is complete at state $\bs_t$ and $0$ otherwise. Formally, the Q-function for instruction $l$ is $Q^\mu_l(\bs_t, \ba_{t:t+H-1}) =
	\mathbb{E}_{\mu}
	\left[
		\sum_{j=0}^{\infty} \gamma^j r_l(\bs_{t+j})|
		\bs_t,\ba_{t:t+H-1}
		\right],$
where $\mu$ denotes the policy used after the initial action chunk and $\gamma \in [0,1)$ is a discount factor.

\textbf{Q-function training approaches.}
We aim to train a generalist Q-function $Q_\theta(\bs_t,\ba_{t:t+H-1},l)$ from an \emph{offline} dataset $\mathcal{D}_\text{pre}$ of robot trajectories. Q-functions can be trained in several ways. The simplest analogue of supervised training is Monte Carlo (MC) regression. Given a trajectory
$\tau \sim \mathcal{D}_\text{pre}$, with
$\tau := (\bs_0,\ba_0,\textbf{r}_0,\bs_1,\ba_1,\ldots)$,
we can compute a return-to-go target from the rewards observed later in the trajectory and regress the Q-function to this target. We call this approach \textbf{MC}. MC regression avoids error compounding from training on self-generated targets; however, the return-to-go target has high variance which affects the quality of the Q-function. Moreover, the learned MC Q-function models the return-to-go under the behavior policy distribution induced by the training dataset, which may be undesirable especially if the data is collected from a suboptimal policy.

An appealing way to address variance is to leverage the Bellman equation and train the Q-function via temporal-difference (TD) learning. Using Q-chunking~\citep{li2025qchunking}, TD learning combines discounted rewards over the action chunk with a \emph{bootstrapped} value estimate from a stale target network $Q_{\bar{\theta}}$:
\begin{align}
	\!\!\!y_t^{\,l}
	 & =
	\sum_{j=0}^{H-1}\gamma^j r_l(\bs_{t+j})
	+
	\gamma^H
	Q_{\bar{\theta}}
	\left(
	\bs_{t+H},
	\ba'_{t+H:t+2H-1},
	l
	\right),~~~
	\mathcal{L}_{\mathrm{TD}}(\theta)
	  =
	\mathbb{E}_{\tau \sim \mathcal{D}_\text{pre},\, t}
	\left[
		\left(
		Q_\theta(\bs_t,\ba_{t:t+H-1},l)
		-
		y_t^{\,l}
		\right)^2
		\right],
	\label{eq:td-learning}
\end{align}
where $\ba'_{t+H:t+2H-1} \sim \mu(\cdot \mid \bs_{t+H})$ is an action chunk used in the backup. The choice of $\mu$ determines the kind of TD update. If $\mu$ is the behavior policy that generated the dataset, then we can simply use the next action chunk $\ba_{t+H:t+2H-1}$ that appears in $\mathcal{D}_\text{pre}$. This approach is called \textbf{SARSA}. Bootstrapping reduces the variance of the target relative to MC, but the bootstrapping errors now compound with every backup, so SARSA does suffer from errors compounding over the horizon. Additionally, since the backup uses the dataset's own actions, the resulting Q-function is fit to the action distribution of the behavior policy.

Alternatively, we could select the action chunk that maximizes the target Q-function at the next state. Computing this maximum exactly is intractable because action chunks are high-dimensional and learning a separate policy-improvement operator is both costly and unstable in offline RL~\citep{mark2025policy}. Instead, we exploit the fact that generalist policies provide a strong prior over the relevant action space: we sample $N$ chunks from the base policy and select the one with the highest target value:
\begin{align}
	\ba^{(i)}_{t+H:t+2H-1}
	\sim
	\pi_\text{base}(\cdot|\bs_{t+H}, l),
	~~~ i=1,\ldots,N, ~~~~~~~~~
	\ba'_{t+H:t+2H-1}
	 & =
	\operatorname*{arg\,max}_{\ba \in \{\ba^{(i)}_{t+H:t+2H-1}\}_{i=1}^N}
	Q_{\bar{\theta}}(\bs_{t+H},\ba,l),
	\label{eq:max-target}
\end{align}
This best-of-$N$ style action selection for the Bellman backup trains the Q-function to evaluate actions that may be better than those observed in the dataset, making it more useful for policy improvement than SARSA. We refer to this approach as TD learning over a best-of-$N$ policy (\textbf{TD-BoN}). Unlike SARSA, however, the backup uses actions sampled from the base policy, so the resulting Q-function is fit to the action distribution of an improved policy. TD-BoN is expected to perform better than SARSA when the training data presents high coverage and several counterfactual (and suboptimal) behaviors. Both approaches that use bootstrapping remain prone to challenges from propagating sparse reward signals over long horizons and from limited coverage of relevant state-action pairs~\citep{park2026horizon}. Modern robotic datasets, which often contain teleoperated or human-recorded long-horizon trajectories, exhibit both challenges. Thus, it is important to carefully design the learning objective for training generalist Q-functions.

\textbf{Language conditioning for generalist Q-functions.}
Independent of the training objective, we must choose what inputs the Q-function obtains. In addition to the state-action pair, we also need to condition any generalist Q-function on some form of a language instruction. There are several choices for this. The first option conditions only on the task instruction $l$, as in $Q_\theta(\bs_t,\ba_{t:t+H-1},l)$ above, so the Q-function models the reward-to-go of the full task. The second option assumes that each trajectory is additionally segmented into subtasks, and denotes the active subtask at state $\bs_t$ by $\tilde{l}_t$. The Q-function then takes both the task and the active subtask as input, $Q_\theta(\bs_t,\ba_{t:t+H-1},l,\tilde{l}_t)$, but is trained to model only the value of the active subtask, i.e., the reward-to-go under the subtask reward $r_{\tilde{l}_t}$ rather than the task reward $r_l$. The task $l$ is included as an input in this option only to make predicting the subtask tractable: the ground-truth subtask $\tilde{l}_t$ is not available at inference time and must be inferred from the observation and the task instruction. Conditioning on the task alone means the Q-function must assign credit over the full long-horizon task, which is precisely what makes value learning struggle in the long-horizon setting.

\textbf{Evaluation setting: policy steering.}
Our downstream evaluation setting is to use a learned Q-function to steer robot behavior when executing policy $\pi_\text{base}$~\citep{nakamoto2025steering}. Ideally, a generalist Q-function should be usable for a new task specified by $l$, either zero-shot or after minimal finetuning to the robot platform, without training a value function from scratch~\citep{intelligence2025pi06vlalearnsexperience}. Given the current state $\bs_t$ and language instruction $l$, the base policy proposes $N$ candidate action chunks. The Q-function scores each chunk, and the robot executes the highest-scoring one, analogous to Equation~\ref{eq:max-target}.
Our problem in this paper is to train a generalist Q-function initialization capable of steering policies across several long-horizon tasks.

\vspace{-0.3cm}
\section{\methodname: Subtask-Elicited Q-Functions for Long-Horizon Manipulation}
\label{sec:method}
\vspace{-0.2cm}

Our aim in this work is to train generalist value functions for long-horizon tasks. This requires addressing two challenges: \textbf{1)} developing a value learning recipe that is effective over long horizons and \textbf{2)} ensuring that this recipe can convert generalist VLMs into effective value functions. In this section, we will develop an approach that addresses both of these challenges.

The central challenge in training Q-functions for long-horizon robotic manipulation is the horizon itself. Standard Q-functions estimate expected success over the entire task, which becomes increasingly difficult as the number of actions grows. Indeed, \citet{park2026horizon} empirically show that Q-estimation error can increase nearly exponentially with horizon in offline RL. The offline setting compounds this problem because datasets provide limited coverage of the many state-action configurations encountered along long trajectories. Thus, before training Q-functions on vision-language model (VLM) backbones, we ask how to reformulate value learning to avoid directly modeling sparse success over the full task horizon.

\begin{figure}[t]
    \centering
    \includegraphics[
        width=0.99\linewidth
    ]{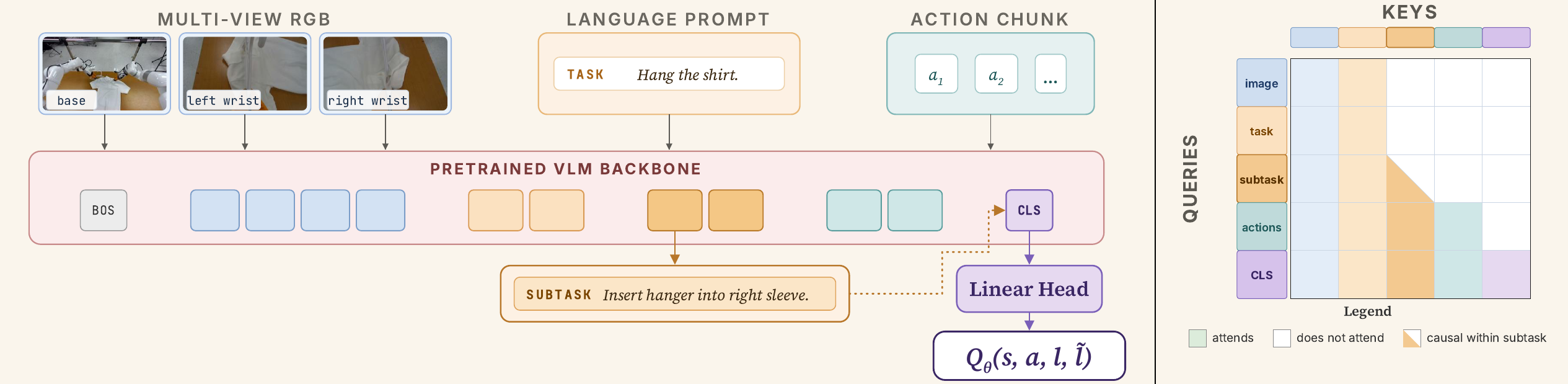}
\caption{\footnotesize{\textbf{Overview of the \methodname{} value-function architecture.}
The model first rolls out the active subtask at state $\bs$ in text using a next-token prediction loss, and then predicts a scalar value $Q_\theta(\bs,\ba,l,\tilde{l})$ for the input state-action pair conditioned on the task instruction and the subtask. Ground-truth subtask annotations supervise the rollout during training; at inference the model conditions on its own predicted subtask.}}
\label{fig:method}
\vspace{-0.2cm}
\end{figure}

\textbf{Key idea: Learning Q-values at the subtask level.}
A natural way to shorten the value-learning horizon is to exploit the local structure of manipulation rollouts, especially when VLM backbones are used for value learning. If a task decomposes into subtask spans that each pursue a local objective, such as an intermediate stage in a multi-stage manipulation task or a new attempt within a precision-heavy stage, the value function can evaluate progress toward the immediate objective rather than predict eventual task success. Classical hierarchical RL uses a related decomposition~\citep{stolle2002learning,nachum2018data}, but typically relies on hand-specified subgoals and a high-level policy that selects among them. In our setting, subtasks are expressed in language, allowing a single language-conditioned Q-function to infer the active subtask and evaluate actions against it.

Following Section~\ref{sec:prelims}, we assume that each trajectory $\tau \in \mathcal{D}_\text{pre}$ is paired with a task instruction $l$ and segmented into subtask spans\footnote{Several robot datasets include subtask annotations. When unavailable, they can be generated by prompting VLMs.}, where $\tilde{l}_t$ denotes the active subtask at $\bs_t$. We define a sparse binary reward for each span, with $r_{\tilde{l}_t}(\bs_t)=1$ when the active subtask terminates and $0$ otherwise, following prior work~\citep{kumar2022pre}. Rather than learning a Q-function for the full task $l$, \methodname{} predicts the value of the input action chunk for the active subtask $\tilde{l}_t$, assuming policy $\mu$ is followed thereafter:\begin{align}
	Q^{\mu}(\bs_t,\ba_{t:t+H-1}, \textcolor{red}{\tilde{l}_t})
	=
	\mathbb{E}_{\mu}
	\Bigg[
		\sum_{j=0}^{\infty}
		\gamma^j \cdot
		r_{\textcolor{red}{\tilde{l}_t}}(\bs_{t+j})
		\Big|\,
		\bs_t,\ba_{t:t+H-1}
		\Bigg].
	\label{eq:subtask-q}
\end{align}
\textbf{Explicitly decoding the subtask.}
Since the ground-truth subtask $\tilde{l}_t$ is not available at inference time, the Q-function must infer it from the observation and the task instruction. \methodname{} explicitly runs this inference: the model first autoregressively decodes the active subtask in text, $\hat{l}_t \sim p_\theta(\cdot \mid \bs_t, l)$, and then predicts the Q-value conditioned on it, $Q_\theta(\bs_t,\ba_{t:t+H-1},l,\hat{l}_t)$. The subtask prediction is supervised with a next-token prediction loss on the ground-truth annotation $\tilde{l}_t = (\tilde{w}_{t,1},\ldots,\tilde{w}_{t,M})$,
\begin{align}
	\mathcal{L}_{\mathrm{subtask}}(\theta)
	=
	-
	\mathbb{E}_{\tau \sim \mathcal{D}_\text{pre},\, t}
	\left[
		\frac{1}{M}
		\sum_{m=1}^{M}
		\log p_\theta
		\left(
		\tilde{w}_{t,m}
		\mid
		\tilde{w}_{t,<m},
		\bs_t,
		l
		\right)
		\right],
	\label{eq:subtask-nll}
\end{align}
while the value head is conditioned on the ground-truth subtask during training. The TD target for Equation~\ref{eq:subtask-q} is constructed using a target network $Q_{\bar{\theta}}$:
\begin{align}
	y_t
	=
	\sum_{j=0}^{H-1}\gamma^j r_{\textcolor{red}{\tilde{l}_t}}(\bs_{t+j})
	+
	\gamma^H\,(1-I_t) \cdot Q_{\bar{\theta}}
	\left(
	\bs_{t+H},
	\ba'_{t+H:t+2H-1},
	l,
	\textcolor{red}{\tilde{l}_{t+H}}
	\right),
	\label{eq:subtask-td-target}
\end{align}
where $\ba'_{t+H:t+2H-1}$ is the backup action chunk and $I_t$ indicates a subtask boundary in $t:t+H-1$. Note that we do not bootstrap across subtask boundaries as it defeats the purpose of subtask-level Q-functions; the backup therefore applies only when $I_t=0$. The overall objective of \methodname{} combines subtask-level TD learning with the subtask-prediction loss:
\begin{align}
	\mathcal{L}_{\methodname{}}(\theta)
	=
	\mathcal{L}_{\mathrm{TD}}(\theta)
	+
	\lambda_{\mathrm{subtask}}
	\mathcal{L}_{\mathrm{subtask}}(\theta),
	\label{eq:seeq-loss}
\end{align}
where $\mathcal{L}_{\mathrm{TD}}$ regresses $Q_\theta$ to $y_t$ as in Equation~\ref{eq:td-learning} and $\lambda_{\mathrm{subtask}}$ controls the strength of the next-token prediction loss.
In principle, the active subtask need not be explicitly decoded: the Q-function could condition only on the full-task instruction while being trained to match the value of the active subtask, relying on its hidden representations to infer that subtask implicitly. However, explicitly decoding the active subtask at inference time and conditioning value prediction is expected to encourage consistency between the inferred subtask and its value. This is particularly important near subtask boundaries, where similar image observations may correspond to different subtasks and implicit inference can produce noisy values that incorrectly steer the policy. \methodname{} therefore decodes the active subtask before predicting its value (Figure~\ref{fig:method}). We visualize value predictions with and without explicit subtask decoding in Figure~\ref{fig:qualitative}.

\textbf{Implementation with a VLM backbone.} We parameterize $Q_\theta$ with a PaliGemma VLM~\citep{beyer2024paligemmaversatile3bvlm} and pretrain it on broad robot data before finetuning on the target task. The VLM provides vision-language priors for recognizing task progress and interpreting instructions, while regularizing these input spaces during downstream finetuning. Because it lacks a prior over robot actions, robot pretraining additionally exposes the model to diverse action chunks across tasks and embodiments, with the aim of regularizing its action conditioning and reducing memorization. We discuss implementation details in Appendix~\ref{app:seeq_architecture}.

While the architecture in \methodname{} can be utilized with any value learning objective, we use the \textbf{TD-BoN objective} as defined in Equation~\ref{eq:max-target}: we sample candidate action chunks from $\pi_\text{base}$ and select the one with the highest value under $Q_{\bar{\theta}}$ as the target action chunk that appears within the Bellman backup. As discussed in Section~\ref{sec:prelims}, this aligns the TD backup with the best-of-$N$ steering procedure used at inference time. We examine value overestimation in Appendix~\ref{app:algorithm}.

\textbf{Test-time inference.} Given a new state $\bs_t$ and task $l$, \methodname{} first rolls out the active subtask $\hat{l}_t$ in text, and then scores each candidate action chunk proposed by $\pi_\text{base}$ with $Q_\theta(\bs_t,\ba_{t:t+H-1},l,\hat{l}_t)$. We use this Q-value for best-of-$N$ policy steering (as discussed in Section~\ref{sec:prelims}).

\vspace{-0.3cm}
\section{Experiments}
\label{sec:experiments}
\vspace{-0.2cm}

\textbf{Implementation setup.} We train \methodname{} in two stages. We first pretrain a PaliGemma-based Q-function on
RoboCOIN~\citep{wu2026robocoinopensourcedbimanualrobotic}, restricted to its bimanual embodiments, which yields 131
tasks and approximately 40k episodes. We then finetune the resulting generalist Q-function on data from each target
task; the amount of downstream data differs across tasks. We use 
$\pi_{0.5}$~\citep{intelligence2025pi05visionlanguageactionmodelopenworld} as our base policy and finetune it directly on the same target-task
data. Across methods, we use $N=8$ candidate action chunks for both TD-BoN backups
during training and best-of-$N$ policy steering at inference. We average the subtask prediction loss over subtask tokens and assign it a weight of $\lambda_\text{subtask} = 0.1$.

\begin{figure}[t]
    \centering
    \captionsetup[subfigure]{font=footnotesize,justification=centering}
    \begin{subfigure}[t]{0.24\linewidth}
        \centering
        \includegraphics[width=\linewidth]{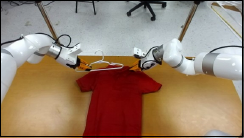}
        \caption{shirt-hang}
    \end{subfigure}\hfill
    \begin{subfigure}[t]{0.24\linewidth}
        \centering
        \includegraphics[width=\linewidth]{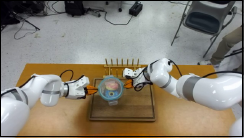}
        \caption{lid-sealing}
    \end{subfigure}\hfill
    \begin{subfigure}[t]{0.24\linewidth}
        \centering
        \includegraphics[width=\linewidth]{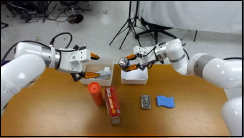}
        \caption{grocery-packing}
    \end{subfigure}\hfill
    \begin{subfigure}[t]{0.24\linewidth}
        \centering
        \includegraphics[width=\linewidth]{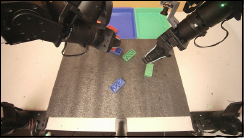}
        \caption{LEGO-disassembly}
    \end{subfigure}
    \vspace{-0.2cm}
    \caption{\footnotesize{\textbf{Our bimanual real-robot evaluation tasks.} Top-camera snapshots from dataset demonstrations of the four real-world tasks. Three tasks are set up on a bimanual xArm-7 platform, and one uses a bimanual YAM platform. See Appendix~\ref{app:eval_protocol} for task definitions and evaluation protocols.}}
    \label{fig:tasks}
    \vspace{-0.4cm}
\end{figure}

\textbf{Tasks.} We evaluate on four real-world, long-horizon bimanual manipulation tasks at 60 Hz (Figure~\ref{fig:tasks}) that test the ability of the Q-function trained via \methodname{} to steer a robot policy through long, multistage tasks that often demand precise low-level behavior. We often observe that a Q-function is needed to select the best action in states requiring precision and to prevent errors from compounding over longer rollouts.
\begin{enumerate}[leftmargin=*,itemsep=1pt,topsep=-2pt]
\item \textbf{shirt-hang}~\citep{hu2025racrobotlearninglonghorizon} (1--2\,min; 386 episodes): remove a hanger from a rod, insert it into both sleeves
of a t-shirt, and hang the shirt back on the rod. The task demands bimanual coordination and precise positioning to
guide the hanger into each sleeve.
\item \textbf{lid-sealing}~\citep{hu2025racrobotlearninglonghorizon} (1--2\,min; 447 episodes): pick up the lid of a food-storage container, place it
on the container, and close all four latching flaps. The task demands precise alignment of the lid and rim for proper sealing.
\item \textbf{grocery-packing}~\citep{anonymous2026vlaexplore} (1--3\,min; 473 episodes): pack deformable grocery items of varying shapes and sizes into boxes. The task demands grasping a diverse set of objects; following instructions, which, unlike in the two tasks above, vary across episodes in both the set of objects and the boxes they are assigned to; and bimanual coordination, as an object must be handed from one arm to the other when it lies on the side opposite its target box.
\item \textbf{LEGO-disassembly} (1--2\,min; 97 training episodes): separate assembled LEGO blocks of different colors
and place each block into the tray of the matching color. Similar to \textbf{grocery-packing}, the instruction varies across
episodes, since the pairing of block colors and tray colors changes. The task demands reasoning about the
orientation in which to hold the blocks and how to pull them apart, as well as bimanual coordination for handoffs.
\end{enumerate}

\begin{figure}[p]
    \centering
    \input{figures/qualitative_strips/figure}
\end{figure}

\textbf{Research questions.} Our experiments are designed to answer the following questions:
\begin{enumerate}[leftmargin=*,itemsep=1pt,label=\textbf{Q\arabic*.}, topsep=-2pt]
\item Can \methodname{} steer the base policy to improve its robustness and performance?
\item What kinds of training data improve the quality of the Q-function learned via \methodname{}?
\item How does \methodname{} compare to standard Monte Carlo/TD-trained value functions at the task level and SARSA-trained value functions at the subtask level?
\item How critical is generalist pretraining to the success of \methodname{}?
\item How much does subtask instruction supervision contribute to the efficacy of \methodname{}, and how important is decoding the predicted subtask at test time?
\end{enumerate}
We will also provide several diagnostic visualizations to understand the behavior of \methodname{}.

\vspace{-0.25cm}
\subsection{Main Results: Policy Steering on Real-World Long-Horizon Tasks}
\vspace{-0.2cm}
\begin{wraptable}{r}{0.45\linewidth}
\centering
\vspace{-0.2cm}
\footnotesize
\setlength{\tabcolsep}{6pt}
\begin{tabularx}{\linewidth}{@{}Xcc|c@{}}
\toprule
\textbf{Task} & \textbf{Data} & \textbf{Base} & \textbf{\methodname{}} \\
\midrule
\textbf{shirt-hang}       & RaC    & 10/24 & \textbf{22/24} \\
\textbf{lid-sealing} & RaC & 10/24 & \textbf{15/24} \\
\textbf{grocery-packing}          & Teleop & 9/24 & \textbf{17/24} \\
\textbf{LEGO-disassembly} & Teleop & 5/24  & \textbf{10/24} \\
\bottomrule
\end{tabularx}
\vspace{-0.1cm}
\caption{\footnotesize{\textbf{Policy steering results.} Success rates over 24 trials for the base policy and for the same policy steered by \methodname{} with best-of-$N$ action selection ($N=8$).}}
\label{tab:main-steering-results}
\vspace{-0.3cm}
\end{wraptable}
We report success rates for the base policy and the same policy steered by a \methodname{} Q-function in Table~\ref{tab:main-steering-results}, selecting among $N=8$ action chunks sampled at each step. We study policy steering under two groups of settings, with datasets that vary in coverage and suboptimality.

\textbf{RaC~\citep{hu2025racrobotlearninglonghorizon} data.} For \textbf{shirt-hang} and \textbf{lid-sealing}, data is collected with RaC~\citep{hu2025racrobotlearninglonghorizon}: the policy runs autonomously until it begins to make a mistake, at which point a human teleoperator takes control to demonstrate corrections and recoveries. These trajectories therefore combine suboptimal policy actions with expert human interventions. When training both the policy and the Q-function on this data, \methodname{} raises the success rate of \textbf{shirt-hang} from 10/24 to 22/24 and of \textbf{lid-sealing} from 10/24 to 15/24.

\textbf{Teleoperation data.} For \textbf{grocery-packing} and \textbf{LEGO-disassembly}, downstream data consists solely of expert demonstrations, collected via human teleoperation. On \textbf{grocery-packing}, \methodname{} improves the success rate of the base policy from 9/24 to 17/24, indicating that steering remains beneficial with expert demonstrations alone. The \textbf{LEGO-disassembly} dataset is particularly limited: it contains only 97 episodes, all starting from the same block configuration (shape and orientation), resulting in narrow state-space coverage. Even in this setting, \methodname{} improves the policy success rate from 5/24 to 10/24.

\vspace{-0.2cm}
\subsection{Comparing \methodname{} to Other Value Learning Objectives}
\label{sec:baselines}
\vspace{-0.1cm}
Next, we compare \methodname{} against baseline and prior approaches for fitting value functions: task-level Monte Carlo return fitting, subtask-level SARSA, and task-level TD learning. We evaluate these methods on \textbf{shirt-hang} and \textbf{grocery-packing}. All methods use the same PaliGemma backbone, are pretrained on RoboCOIN, and are finetuned on identical task-specific data. They differ only in the return they predict, the action used for bootstrapping, and whether they explicitly predict the active subtask. The task-level methods use a discount factor of $0.9995$, compared with $0.999$ for the subtask-level methods, to accommodate the longer return horizon.

\begin{wraptable}{r}{0.4\linewidth}
\centering
\vspace{-0.4cm}
\footnotesize
\setlength{\tabcolsep}{3pt}
\makebox[\linewidth][l]{\begin{tabular}{@{}lcc@{}}
\toprule
\textbf{Objective} & \textbf{shirt-hang} & \textbf{grocery-packing} \\
\midrule
Base policy        & 10/24          & 9/24 \\
\noalign{\vskip 2pt}
\hdashline
\noalign{\vskip 2pt}
\textit{Task-level} MC      & 9/24           & 10/24 \\
\textit{Task-level} TD  & 12/24          & 9/24 \\
\noalign{\vskip 2pt}
\hdashline
\noalign{\vskip 2pt}
\textit{Subtask-level} SARSA & 13/24          & 10/24 \\
\noalign{\vskip 2pt}
\hdashline
\noalign{\vskip 2pt}
\textbf{\methodname{}} (Ours)      & \textbf{22/24} & \textbf{17/24} \\
\bottomrule
\end{tabular}}
\vspace{-0.2cm}
\caption{\footnotesize{\textbf{Comparison of value-learning objectives.}
Success rates on \textbf{shirt-hang} and \textbf{grocery-packing}. Fitting values of the data collection policy underperforms \methodname{}, while task-level MC and TD offer little improvement over the base policy.}}
\label{tab:objective-ablation}
\vspace{-0.4cm}
\end{wraptable}
\textbf{Task-level MC.} Prior work learns state-based value functions by regressing to observed MC returns~\citep{fei2026wcm,intelligence2025pi06vlalearnsexperience}. We apply this strategy to an action-value function to test whether direct return regression suffices for policy steering, compared with \methodname{}'s subtask-based TD approach. This baseline conditions on the full-task instruction $l$ and fits the observed discounted return to task completion, as described in Section~\ref{sec:prelims}. Concretely, it uses the squared-error loss in Equation~\ref{eq:td-learning} with $y_t^{\,l}$ replaced by $G_t^l=\sum_{j=0}^{T-t}\gamma^j r_l(\bs_{t+j})$, where $T$ is the final timestep of the recorded trajectory. We also tried alternative losses, including cross-entropy~\citep{farebrother2024stop}, in preliminary experiments but found no noticeable difference in performance. The targets are computed entirely from the dataset, with no bootstrapping or target network. This baseline neither predicts nor conditions on a subtask and uses no next-token prediction loss.

\textbf{Subtask-level SARSA.} This baseline retains \methodname{}'s subtask rewards, boundary termination, autoregressive subtask conditioning, and next-token loss. It replaces the backup action in Equation~\ref{eq:subtask-td-target} with the next action chunk recorded in the dataset, $\ba'_{t+H:t+2H-1}=\ba_{t+H:t+2H-1}$, following the SARSA update in Equation~\ref{eq:td-learning}. The value loss therefore evaluates continuation under the dataset's behavior policy, without maximizing over policy candidates as in Equation~\ref{eq:max-target}. Comparing this baseline with \methodname{} tests whether the best-of-$N$ backup improves steering beyond subtask-level policy evaluation.

\textbf{Task-level TD learning.} This baseline tests whether a best-of-$N$ TD objective is effective without the subtask formulation. It minimizes the TD loss in Equation~\ref{eq:td-learning} using the full-task reward $r_l$ and the best-of-$N$ backup in Equation~\ref{eq:max-target}. Like task-level MC, it conditions only on $l$ and uses neither subtask prediction nor next-token loss. Its backups continue across intermediate subtask boundaries and terminate only at the end of the task. All TD variants use multi-step targets that accumulate discounted rewards over the backup interval.

\textbf{Results.} Table~\ref{tab:objective-ablation} shows that \methodname{} outperforms all evaluated baselines on both tasks. Task-level MC and TD learning offer little improvement over the base policy, while subtask-level SARSA provides a modest gain on \textbf{shirt-hang} and a one-trial gain on \textbf{grocery-packing}. These results suggest that both the subtask formulation and the best-of-$N$ backup contribute to effective policy steering, with gains on both mixed policy and human intervention data and purely expert demonstrations.

\vspace{-0.2cm}
\subsection{Importance of Pretraining and VLM Initialization for \methodname{}}
\label{sec:pretraining}
\vspace{-0.1cm}
To quantify the benefits of initializing \methodname{} from a general VLM backbone and pretraining on broad robot data, we evaluate two baselines. First, we compare against a task-specific value function trained on target-task data alone. For a fair comparison, this baseline uses the same TD-learning objective and best-of-$N$ policy extraction with the same base policy. It also predicts the active subtask as a categorical token and conditions its value prediction on this token. Its architecture consists of a pretrained ResNet-50 image encoder followed by an MLP, following the design in \citet{kumar2022pre}. Each camera's feature map is pooled using learned spatial weights and projected to an embedding. The value MLP combines these camera embeddings with the flattened action chunk and a learned subtask embedding.
Next, we evaluate \methodname{} without robot data pretraining. This ablation retains \methodname{}'s architecture and training objective but skips the robot data pretraining stage, finetuning the pretrained PaliGemma backbone directly on the target task, \textbf{shirt-hang}. We observe in Table~\ref{tab:task-specific} that with the same VLM backbone and \methodname{} objective, general robot data pretraining raises success from 8/24 to 22/24. In addition, \methodname{} outperforms a task-specific value function, indicating that both pretraining and VLM initialization are critical.
\begin{table}[H]
\centering
\footnotesize
\setlength{\tabcolsep}{6pt}
\begin{tabular}{@{}lccc@{}}
\toprule
\textbf{Value function} &
\textbf{Robot pretraining} &
\textbf{VLM initialization} &
\textbf{Success rate} \\
\midrule
Base policy & --- & --- & 10/24 \\
\noalign{\vskip 2pt}
\hdashline
\noalign{\vskip 2pt}
Task-specific (ResNet-50 + MLP, TD-BoN)
& \ding{55} & \ding{55} & 16/24 \\
\methodname{} minus pretraining
& \ding{55} & \ding{51} & 8/24 \\
\noalign{\vskip 2pt}
\hdashline
\noalign{\vskip 2pt}
\textbf{\methodname{} (Ours)}
& \ding{51} & \ding{51} & \textbf{22/24} \\
\bottomrule
\end{tabular}
\vspace{-0.2cm}
\caption{\footnotesize{\textbf{Importance of pretraining and VLM initializations for \methodname{}, evaluated on \textbf{shirt-hang}.}
Columns indicate whether the value function uses robot data pretraining and VLM initialization. Observe that both robot data pretraining and base VLM initialization are important for the success of \methodname{}.}}
\label{tab:task-specific}
\vspace{-0.4cm}
\end{table}

We hypothesize that \methodname{} without robot pretraining performs poorly
because robot actions come from a distribution unseen during
pretraining of the VLM backbone. Direct target-task finetuning may therefore
overfit to image features, while pretraining on diverse robot data can help
alleviate this issue. Figure~\ref{fig:generalist-smoothness}(b) compares image-gradient
norms with and without robot pretraining to examine this hypothesis and provides evidence that supports this hypothesis: note that the sensitivity of the Q-function to visual observations is higher without pretraining\footnote{The sensitivity to \emph{both} images and actions is higher for the ResNet-50 baseline, implying that it might have spuriously fit to both action and state features, while no robot pretraining with \methodname{} suppresses sensitivity to actions specifically.}.

\vspace{-0.2cm}
\subsection{Ablation Study: Effect of Predicting Subtasks and Conditioning on Them}
\label{sec:subtask-ablation}
\vspace{-0.2cm}
Next, we assess the importance of predicting the active subtask and conditioning value estimation on it for \textbf{shirt-hang}. In the first variant, we remove subtask prediction and conditioning, training the model with the subtask-level TD loss alone. In the second variant, we retain the subtask prediction loss but do not condition the value head on the subtask. The subtask is therefore not decoded at inference time and is represented only implicitly in the model's activations.

\begin{wraptable}{r}{0.55\linewidth}
\centering
\footnotesize
\setlength{\tabcolsep}{4pt}
\begin{tabular}{@{}p{0.70\linewidth}c@{}}
\toprule
\textbf{Variant} & \textbf{Success rate} \\
\midrule
Base policy & 10/24 \\
\noalign{\vskip 2pt}
\hdashline
\noalign{\vskip 2pt}
Subtask-level TD\newline
{\scriptsize\textit{No subtask prediction, no subtask conditioning}} & 8/24 \\
\noalign{\vskip 2pt}
\hdashline
\noalign{\vskip 2pt}
Subtask-level TD\newline
{\scriptsize\textit{Subtask prediction, no subtask conditioning}} & 15/24 \\
\noalign{\vskip 2pt}
\hdashline
\noalign{\vskip 2pt}
\textbf{\methodname{}} (Ours) & \textbf{22/24} \\
\bottomrule
\end{tabular}
\vspace{-0.2cm}
\caption{\footnotesize{\textbf{Ablation of subtask elicitation on \textbf{shirt-hang}.}
Subtask prediction supervision improves success, with further gains from explicitly conditioning value estimation on the predicted subtask.}}
\label{tab:ntp-ablation}
\vspace{-0.6cm}
\end{wraptable}
As shown in Table~\ref{tab:ntp-ablation}, removing the subtask prediction loss causes the steered policy to underperform the base policy (8/24 vs.\ 10/24). Adding the prediction loss improves success to 15/24, even without conditioning value predictions on the predicted subtask. Explicitly decoding the subtask and conditioning value estimation on it further improves success to 22/24. Figure~\ref{fig:qualitative} illustrates how subtask conditioning aligns value resets with predicted subtask changes.

\vspace{-0.2cm}
\subsection{Diagnostic Visualization: Qualitative Analysis of Value-Function Landscapes}
\label{sec:qualitative-smoothness}
\vspace{-0.1cm}
VLM pretraining provides regularization in the vision-language space. We hypothesize that pretraining on diverse robot data helps the critic learn action conditioning while reducing overfitting to image features during target-task finetuning. To examine this hypothesis, we compare \methodname{}, \methodname{} without robot pretraining, and the task-specific ResNet-50 critic on six complete held-out \textbf{shirt-hang} trajectories. All three critics use subtask-level returns and condition on their own predicted subtask, decoded at every frame. At each frame, they score the same eight cached $\pi_{0.5}$ action chunks, and both gradients are evaluated at that critic's highest-valued candidate. The selected action can differ across critics.

\begin{figure}[t]
    \centering
    \includegraphics[width=0.49\linewidth]{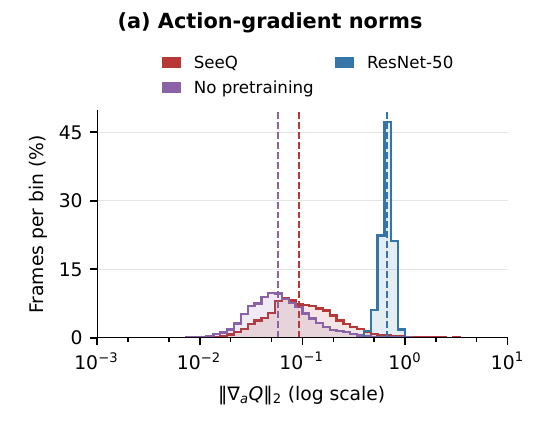}\hfill
    \includegraphics[width=0.49\linewidth]{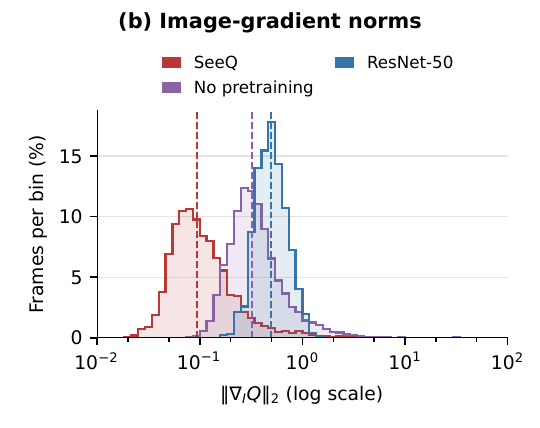}
    \vspace{-0.3cm}
    \caption{\footnotesize{\textbf{Action and image sensitivity on six complete held-out \textbf{shirt-hang} trajectories.}
    (a) Action-gradient norms. (b) Joint image-gradient norms across three cameras.
    All three critics use the same $12{,}582$ frames and eight cached policy candidates;
    gradients are evaluated at each critic's selected candidate. No pretraining denotes
    \methodname{} finetuned directly from PaliGemma. Bar heights are percentages of frames in shared logarithmically spaced bins; dashed lines denote medians.}}
    \label{fig:generalist-smoothness}
    \vspace{-0.3cm}
\end{figure}

Figure~\ref{fig:generalist-smoothness} compares local sensitivity to actions and images. Panel (a) shows $\|\nabla_{\ba}Q\|_2$ in normalized $60\times14$ action coordinates. Panel (b) shows $\|\nabla_I Q\|_2$, with $I$ concatenating all three $224\times224$ RGB camera inputs in the network's $[-1,1]$ coordinates. Both histograms pool all $12{,}582$ frames.

\textbf{Results.} We observe that \methodname{} attains a median image-gradient norm of $0.093$, compared
with $0.323$ without robot pretraining and $0.494$ for ResNet-50.
The $3.5\times$ reduction relative to direct VLM finetuning is consistent with our
hypothesis that robot pretraining reduces sensitivity to image features.
Its median action-gradient norm is $0.093$, compared with $0.673$ for ResNet-50
($7.2\times$ lower). However, the no-pretraining critic has an even smaller
action-gradient median of $0.058$ despite worse steering performance (8/24),
suggesting that it may be insufficiently discriminative between candidate actions
and may not use the action input meaningfully.

\vspace{-0.2cm}
\section{Related Work}
\label{sec:citations}
\vspace{-0.2cm}

Recent robot learning has increasingly followed the recipe of scaling data and model capacity, with substantial success~\citep{zitkovich2023rt,kimopenvla,o2024open,intelligence2025pi05visionlanguageactionmodelopenworld}. However, pure imitation learning struggles to learn from diverse, potentially suboptimal data~\citep{kumar2021should} and remains brittle on long-horizon tasks, where errors compound over time~\citep{ross2010efficient,belkhale2023hydra}. To address these limitations, prior work has explored training value functions with RL~\citep{kumar2022pre,bhateja2023robotic} and using them to guide~\citep{nakamoto2025steering} or improve policies~\citep{mark2025policy}. However, these methods typically instantiate value functions with small networks and use little pretraining data~\citep{bhateja2023robotic,kumar2022pre,yang2024robot,nakamoto2025steering}, if any~\citep{mark2025policy}. Thus, the value functions do not inherit the broad pretraining that makes generalist policies useful, and hence, require task-specific demonstrations before they can be deployed.

A recent line of work has sought to scale value-function learning in robotics in the same spirit as generalist imitation learning~\citep{springenberg2024offline,chebotar2023q,mavip,lee2026roborewardgeneralpurposevisionlanguagereward,kim2026cosmos}. For example, \citet{springenberg2024offline} and \citet{chebotar2023q} train large transformer backbones from scratch with temporal-difference (TD) learning, but do not leverage modern large-scale vision-language pretraining. In contrast, \citet{lee2026roborewardgeneralpurposevisionlanguagereward} finetune a Qwen-3-VL model~\citep{bai2025qwen3} on broad robot data to obtain a VLM-scale reward model. Related efforts learn purely state-based value or progress estimates: WCM~\citep{fei2026wcm} regresses Monte Carlo (MC) returns, RynnValue~\citep{huang2026rynnvalue} predicts observed time-to-completion, and Robometer~\citep{liang2026robometer} combines frame-level progress and success supervision with trajectory preferences. These predictions are trained using trajectory-derived targets rather than TD backups. Such approaches address the challenge of fitting value or reward models at large parametric scale, but long-horizon value learning remains difficult. Direct supervision with MC returns~\citep{kim2026cosmos,fei2026wcm} or observed progress avoids TD credit assignment, but ties the targets to the behaviors and coverage of the training data. Conversely, methods that rely on TD learning~\citep{wang2026learningdeployingfleetscalereinforcement} must propagate sparse success signals over long horizons. Our work addresses this tradeoff by training an action-value function with TD over shorter-horizon subtasks, supporting test-time steering against the learned critic. We further show that explicitly predicting the subtask in text improves value estimation, as language can supervise intermediate features for assessing the active subtask at a given state.

Prior work has also explored similar styles of automatic subtask decomposition, but primarily for policy learning rather than value learning~\citep{intelligence2025pi05visionlanguageactionmodelopenworld,zhang2024universal,ahn2022can}. For example, \citet{intelligence2025pi05visionlanguageactionmodelopenworld} train a high-level VLM to predict per-step subtask instructions for a low-level controller, while \citet{zhang2024universal} discover visual subgoals from phase shifts in a pretrained representation and use them to condition imitation policies and shape rewards. In contrast, we use subtask decomposition to define the prediction problem for a generalist Q-function: the model infers the current subtask and estimates the value of an action for completing that subtask at test time, without requiring subtask annotations or any additional subtask information at deployment. This makes our approach effective yet free of test-time assumptions.

\vspace{-0.3cm}
\section{Discussion, Conclusion, and Future Work}
\vspace{-0.1cm}

We presented \methodname{}, an approach for training generalist Q-functions that steer robot policies on long-horizon tasks. By predicting values for the active subtask, \methodname{} shortens the credit-assignment horizon while retaining TD learning and best-of-$N$ backups for policy improvement. Explicitly predicting the subtask in language allows the value function to use this structure without requiring subtask annotations at deployment. Across four real-world bimanual manipulation tasks, \methodname{} raises the average success rate from 35.4\% to 66.7\% ($1.88\times$), with improvements on various data compositions. Our ablations highlight the importance of robot data pretraining, the best-of-$N$ backup, and explicit subtask conditioning. These findings suggest that choosing an appropriate prediction horizon and using language to structure value estimation are useful ingredients for generalist value learning.

\textbf{Limitations and future work.} \methodname{} relies on subtask annotations during training, but decomposing manipulation trajectories into discrete spans can be ambiguous, particularly during recoveries or near subtask boundaries. Incorporating past frames and using finer-grained annotations may partly mitigate this ambiguity, but the appropriate subtask granularity remains task-dependent. At deployment, errors in subtask prediction can cause the critic to evaluate actions against an incorrect objective. Learning decompositions that are useful for value estimation and accounting for uncertainty over the active subtask are promising directions for addressing this limitation. Moreover, optimizing the value of the current subtask favors local progress, which need not align with the overall task. For example, a robot may knock over another object while packing the current one, making subsequent subtasks harder without reducing progress on the active subtask. Combining subtask-level values with estimates of downstream consequences could address this tradeoff while preserving the benefits of shorter-horizon learning.

Finally, our evaluation uses the Q-function to select among action chunks proposed by a fixed base policy, so steering is limited by the quality and diversity of these candidates. Producing more exploratory base policies and using the learned Q-function to directly improve the policy is a natural next step. Our experiments also finetune the generalist value function on each target task; evaluating transfer with less downstream data, or without task-specific finetuning, would further clarify the scope of its generalization.

\vspace{-0.2cm}
\section*{Acknowledgements}
\vspace{-0.2cm}
We thank Yizhou Li for her help with subtask annotation of the existing \textbf{lid-sealing} RaC data and data collection for the \textbf{LEGO-disassembly} task.
We thank Kshitiz and Robyn Wu for support with the bimanual grocery-packing setup and data from their forthcoming work~\citep{anonymous2026vlaexplore}. We thank Niharika Pant and Naveen Enock for help with robot setups. We thank Lehong Wu, Anthony Liang, Abhishek Gupta, Aykut Onol, and Kushal Arora for informative discussions and feedback on an earlier version of this work. We thank members of the CMU AIRe lab for their support.

This work is primarily supported by a Toyota Research Institute U3.0 award. We also acknowledge support from the  Office of Naval Research under N00014-24-12206 and a Google TPU Builders program gift. We thank the TPU research cloud (TRC) program for their support with Google TPU resources that made this work possible and Gemini Academic Grants program for providing Gemini credits.

\bibliography{main}  %

\newpage
\appendix
\onecolumn

\section{Appendices}

\subsection{Additional Implementation Details}
\label{app:exp_details}

\subsubsection{Action Space and Normalization}
\label{app:action_space}
Value-function pretraining and finetuning use a shared $14$-dimensional bimanual
end-effector (EEF) action space. The base policies use the same representation except
on \textbf{LEGO-disassembly} (YAM platform), where the policy operates in joint space. For each of
the two arms, an EEF action contains
translational deltas $(\Delta x,\Delta y,\Delta z)$, roll--pitch--yaw orientation deltas,
and a gripper command, giving the layout
\[
\big[\underbrace{\Delta x,\Delta y,\Delta z}_{\text{left pos}},\;
\underbrace{\Delta \phi,\Delta \theta,\Delta \psi}_{\text{left rot}},\; g_{\text{left}},\;
\underbrace{\Delta x,\Delta y,\Delta z}_{\text{right pos}},\;
\underbrace{\Delta \phi,\Delta \theta,\Delta \psi}_{\text{right rot}},\; g_{\text{right}}\big].
\]
We use the chunk-wise delta parameterization of~\citet{feng2026demystifyingactionspacedesign}:
every action in a chunk is expressed relative to the current state rather than
recursively relative to the previous predicted action, with orientation deltas composed
as relative rotations on the rpy slots. Deltas are applied to all non-gripper dimensions;
the two gripper channels are absolute. The proprioceptive state $\bs^p_t$ uses the same
$14$-D EEF layout.

For \textbf{LEGO-disassembly}, demonstrations are collected using a bimanual YAM station and
recorded as six joint-angle targets and a gripper command per arm, matching the
joint-position interface used by our teleoperation stack. The base policy is trained
to predict joint-angle deltas relative to the current joint configuration, with
absolute gripper commands. To retain the EEF representation used during
\methodname{} pretraining, we apply a computationally inexpensive forward-kinematics
(FK) transform to the recorded joint targets and current joint configuration, then
express the resulting target poses as EEF deltas relative to the current pose for
critic training. At inference, the same conversion maps each joint-space policy
candidate into EEF space for critic scoring; the robot executes the selected
candidate in joint space.

Both state and action chunks are normalized with quantile normalization: each
dimension is mapped via its $1^{\text{st}}$/$99^{\text{th}}$ percentiles,
$\tilde{x} = 2\,\tfrac{x-q_{01}}{q_{99}-q_{01}}-1$, and then clipped to
$[-1.25,\,1.25]$. The same normalize-then-clip scheme is applied to states, action
chunks and the action candidates used in the TD backup. Quantile normalization and
clipping are common to value-function pretraining, finetuning, and base-policy
training, with statistics computed in the representation used by each model. For
\textbf{LEGO-disassembly}, we undo the policy's normalization and recover absolute
joint targets before applying FK; the resulting EEF deltas are then normalized and
clipped using the critic's statistics before scoring.

\subsubsection{Handling Mixed Control Frequencies in the RoboCOIN Dataset}
\label{app:control_rate}
Our pretraining dataset, RoboCOIN, aggregates demonstrations collected at two different control rates: part of the
data is recorded at $30$ fps and part at $50$ fps. Rather than resampling the
trajectories, the dataloader places both rates on a common wall-clock timeline, so that a single discount factor, a single TD horizon, and
a single action-chunk length all correspond to the same real-time duration regardless of
the source rate. In effect, the per-frame discount is applied per unit of time rather
than per frame, and the $H$-step lookahead used to form the bootstrap target is expressed
as a fixed time window (e.g.\ $1$\,s) that spans proportionally more frames in the
$50$ fps data than in the $30$ fps data.

This rate normalization is what keeps the value targets consistent across the two sources.
The MC return to the end of the active subtask, the $H$-step reward, and the
termination flag are all computed on this normalized timeline: the subtask-completion
reward fires, and the transition is marked terminal, when the active subtask ends
within the TD window, with the reward discounted to the subtask boundary; otherwise
the reward is zero and the value bootstraps from the next state using a discount matched to
the same time window. Because a fixed-length action chunk covers more wall-clock time at a
lower frame rate, the $30$ fps trajectories additionally use only the leading portion of
each chunk (the trailing slots are masked), so the effective action horizon again matches
across the two rates.

\subsubsection{Training and Inference Algorithm Details}
\label{app:algorithm}
Algorithm~\ref{alg:seeq} summarizes the \methodname{} training procedure used for both
pretraining and finetuning, following Section~\ref{sec:method}.
Unlike the per-recorded-step discount in the main text, the algorithm's $\gamma$
is defined at the reference rate $c=150$\,Hz. At recording rate $f$, the main-text
discount therefore corresponds to $\gamma^{c/f}$ here, and a $T$-second backup
uses $\gamma^{cT}$.
The model predicts
the active subtask autoregressively from the visual observation and task instruction,
then evaluates the action chunk conditioned on that subtask. During training,
ground-truth subtask tokens provide teacher-forced supervision for the next-token
loss and remain visible to the action tokens and the value token.
Causal attention within the subtask sequence ensures that each token is predicted
from the observation, task instruction, and preceding subtask tokens.

The rewards and discounts in the TD targets follow the control-rate normalization
in Appendix~\ref{app:control_rate}. In the algorithm, $r_{\tilde{l}_t}(\bs_t)$ denotes
the discounted subtask-completion reward within the TD window, and bootstrapping
stops when the active subtask terminates within that window.
Both current and target Q-values are conditioned on the corresponding
ground-truth subtask during training. Base-policy candidates for the TD backups are
sampled and cached offline.

\begin{algorithm}[t]
\caption{Training \methodname{}}
\label{alg:seeq}
\begin{algorithmic}[1]
\REQUIRE subtask-annotated dataset $\mathcal{D}$; base policy $\pi_\text{base}$;
discount $\gamma$; learning rate $\eta$; target EMA rate $\tau$;
chunk duration $T$ (wall-clock time spanned by one action chunk);
action horizon $H=\mathrm{round}(f_{\mathrm{fps}}\,T)$ (chunk length at the recording frame rate);
reference control rate $c=150$\,Hz (Appendix~\ref{app:control_rate});
subtask-loss weight $\lambda_{\mathrm{subtask}}=0.1$; backup width $N=8$
\STATE Initialize $\theta$ from PaliGemma for pretraining, or from pretrained \methodname{} for finetuning
\STATE Initialize target parameters $\bar{\theta}\leftarrow\theta$
\FOR{each training step}
    \STATE Sample a minibatch $\mathcal{B}$ of annotated transitions
    $(\bs_t,\ba_{t:t+H-1},r_{\tilde{l}_t},\bs_{t+H},l,\tilde{l}_t,\tilde{l}_{t+H})$
    \STATE For each transition, retrieve $N$ cached chunks $\ba^{(i)}_{t+H:t+2H-1}\sim\pi_\text{base}(\cdot\mid\bs_{t+H},l)$, $i=1,\dots,N$
    \STATE $\ba'_{t+H:t+2H-1}\leftarrow\operatorname*{arg\,max}_{\ba\in\{\ba^{(i)}\}_{i=1}^N}\,Q_{\bar{\theta}}(\bs_{t+H},\ba,l,\tilde{l}_{t+H})$
    \STATE $y_t\leftarrow r_{\tilde{l}_t}(\bs_t)+\gamma^{c\,T}\,(1-I_t)\,Q_{\bar{\theta}}(\bs_{t+H},\ba'_{t+H:t+2H-1},l,\tilde{l}_{t+H})$ \COMMENT{$I_t$: subtask boundary in $t:t+H-1$}
    \STATE $\mathcal{L}_{\mathrm{TD}}\leftarrow\mathbb{E}_{\mathcal{B}}\big[\big(Q_\theta(\bs_t,\ba_{t:t+H-1},l,\tilde{l}_t)-\mathrm{sg}(y_t)\big)^2\big]$ \COMMENT{$\mathrm{sg}$: stop-gradient}
    \STATE Compute $\mathcal{L}_{\mathrm{subtask}}$ by teacher forcing $\tilde{l}_t$ (Eq.~\ref{eq:subtask-nll}), averaging over tokens within each example and then over $\mathcal{B}$
    \STATE $\mathcal{L}\leftarrow\mathcal{L}_{\mathrm{TD}}+\lambda_{\mathrm{subtask}}\mathcal{L}_{\mathrm{subtask}}$
    \STATE Update $\theta$ with AdamW using $\nabla_\theta\mathcal{L}$ and learning rate $\eta$
    \STATE $\bar{\theta}\leftarrow(1-\tau)\,\bar{\theta}+\tau\,\theta$
\ENDFOR
\end{algorithmic}
\end{algorithm}

At inference, given $(\bs_t,l)$, \methodname{} first autoregressively decodes the active
subtask $\hat{l}_t$. The base policy proposes $N=8$ action chunks, and the critic scores
each candidate as $Q_\theta(\bs_t,\ba^{(i)},l,\hat{l}_t)$. The robot executes the
highest-scoring candidate, using the predicted subtask in place of the ground-truth
annotation supplied during training. No external subtask annotations or manual
subtask switching are required at inference.

\begin{figure}[t]
    \centering
    \includegraphics[width=0.49\linewidth]{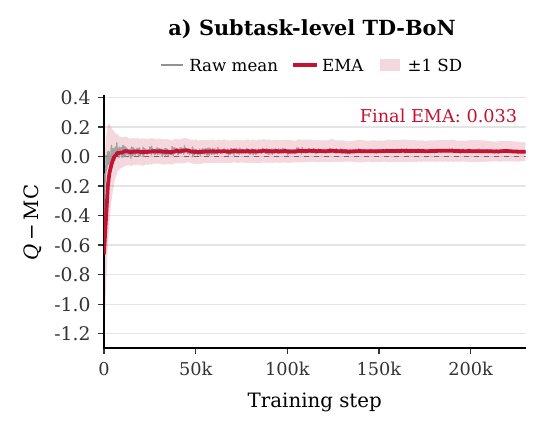}\hfill
    \includegraphics[width=0.49\linewidth]{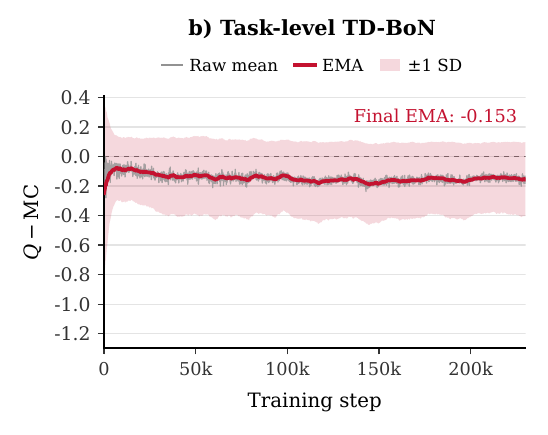}
    \caption{\footnotesize{\textbf{Q-values relative to recorded returns during RoboCOIN pretraining.}
    Gray: batch-mean $Q-\mathrm{MC}$; red: EMA with a $1{,}000$-step half-life;
    shading: $\pm1$ standard deviation of the exponentially weighted prediction-minus-return
    differences.
    (a) \methodname{} with subtask returns shows a near-zero gap after initial underestimation.
    (b) Task-level TD-BoN persistently underestimates task returns.}}
    \label{fig:value-overestimation}
\end{figure}

\textbf{Value overestimation.}
We monitor the difference between predicted Q-values for dataset action chunks and
their recorded MC returns during RoboCOIN pretraining, using subtask returns for
\methodname{} and task returns for task-level TD-BoN. The expected
$Q-\mathrm{MC}$ gap upper-bounds mean signed overestimation relative to the optimal
value for the corresponding reward: an optimal continuation achieves at least the expected return of
the recorded behavior. A positive gap can therefore reflect both estimation error
and improvement over the recorded continuation. This interpretation applies in
expectation on dataset actions. Figure~\ref{fig:value-overestimation}(a) shows this gap
for \methodname{} over the full pretraining run.

\subsubsection{\methodname{} Architecture}
\label{app:seeq_architecture}
Figure~\ref{fig:method} summarizes the PaliGemma-based architecture of \methodname{}.
SigLIP encodes the camera images into patch tokens; Gemma processes these with
task and subtask text, linearly projected action tokens, and a learned value token
at the end of the sequence. Image and task tokens attend bidirectionally.
Subtask tokens use a causal mask: they attend to the images, task instruction,
and preceding subtask text, but not to later actions or the value token.
Action tokens attend bidirectionally within the candidate chunk and can attend
to all preceding image and language tokens, including the subtask.

The final value token attends to all valid preceding tokens, combining the
observed state, language context, and candidate actions; a linear readout maps its
final representation to a scalar Q-value. Earlier tokens cannot attend to the
value token. Padding and unused action positions are masked. We do not use
proprioceptive state as an input to any of the value functions trained. During training, the model
receives ground-truth subtask tokens; at inference, it first decodes the subtask
and then scores candidate action chunks.

\subsubsection{Evaluation Protocol}
\label{app:eval_protocol}
For each of the four tasks, we evaluate the models being compared over $24$ trials.
Randomization is consistent across models: trial $i$ uses the same physical setup and
task instruction for every model evaluated on that task. Figure~\ref{fig:tasks}
illustrates the four tasks using top-camera snapshots from dataset demonstrations.
In all evaluations, each action chunk spans $1$\,s, and actions are replanned every $0.5$\,s.
For evaluations that infer the active subtask and condition value predictions on it, the subtask is decoded every four replanning calls ($2$\,s).
\begin{enumerate}[leftmargin=*,itemsep=2pt]
\item \textbf{shirt-hang}. We randomize the shirt's position and tilt across trials.
The hanger's position also varies, and the robot must insert the hanger into both
sleeves of the shirt and hang it on the rod.
\item \textbf{lid-sealing}. We randomize the positions of the
container and the lid on the dish rack. The robot must pick up the lid and align it
with the container. Successful sealing requires closing all four latching flaps.
\item \textbf{grocery-packing}. Each trial uses two of three box types, with eight trials
for each of the three possible pairs. Each trial has a different language instruction
specifying the set of objects to pack. The exact object sets here are absent from the training
distribution, testing generalization to new packing requests.
\item \textbf{LEGO-disassembly}. The assembled blocks have the same initial shape
and orientation as in training. We randomize the language instruction specifying
the mapping from block colors to tray colors. The robot must disassemble the blocks
and place them in the trays according to this mapping.
\end{enumerate}

\subsection{Additional Experimental Results}
\label{sec:addl_results}

\subsubsection{Subtask Prediction and Value Conditioning}
\label{app:subtask-conditioning}
The ablation in Section~\ref{sec:subtask-ablation} separates the benefit of the
auxiliary next-token prediction (NTP) loss from that of explicitly conditioning
values on the decoded subtask. Skipping subtask decoding is an appealing way to
reduce inference latency: joint training might allow latent activations to carry
the relevant subtask information implicitly. A similar motivation appears in
video--action models such as UVA and Fast-WAM, which retain video supervision
while bypassing explicit video generation at action-inference time
\citep{li2025unifiedvideoactionmodel,yuan2026fastwam}.

We compare \methodname{} with the variant that retains the auxiliary NTP loss but
does not condition its value head on subtask tokens. Both use subtask-level TD-BoN
and the same NTP weight of $0.1$, and are finetuned on
\textbf{shirt-hang} with the same learning-rate schedule, starting from their
corresponding RoboCOIN-pretrained checkpoints.

\begin{figure}[t]
    \centering
    \includegraphics[width=\linewidth]{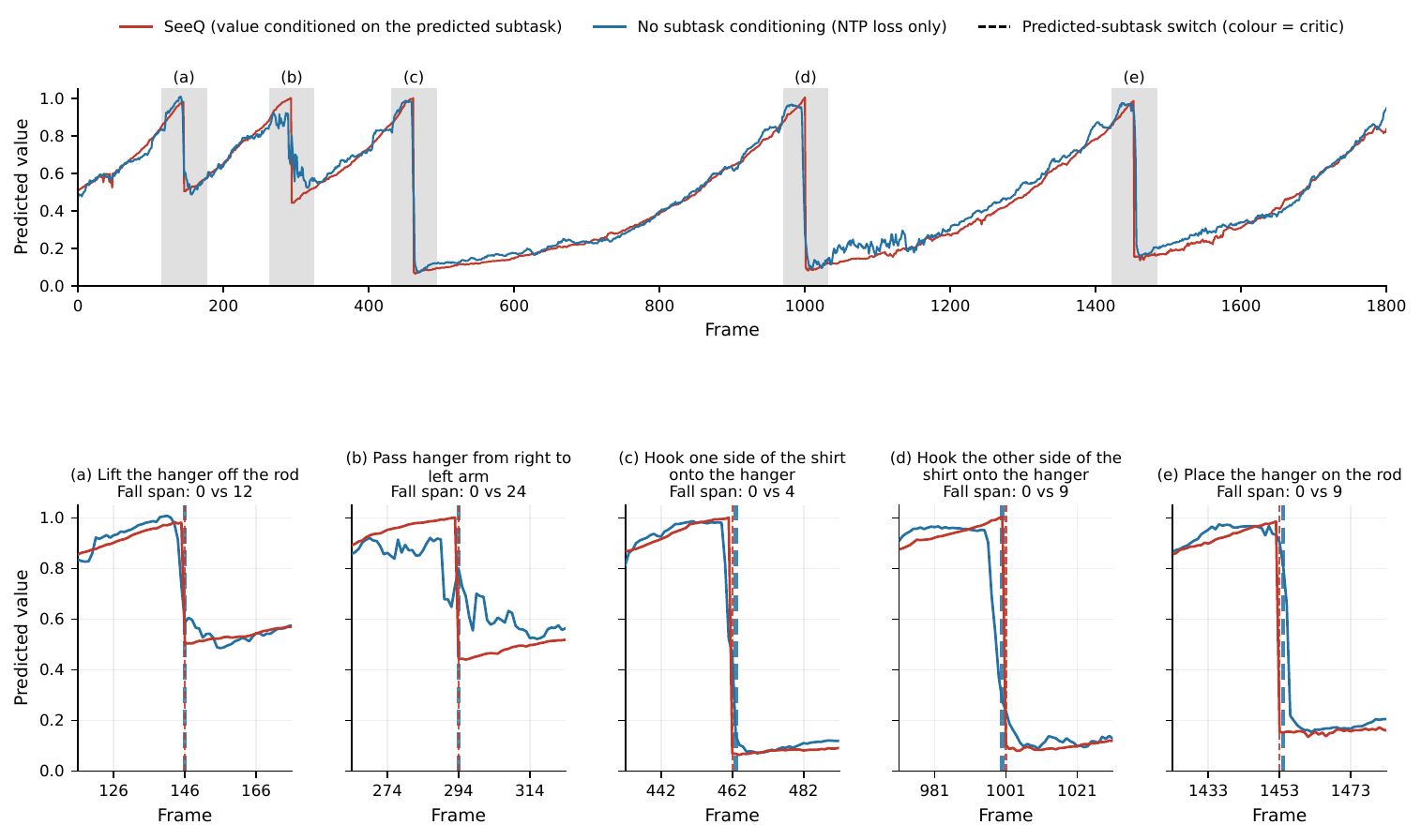}
    \caption{\footnotesize{\textbf{Subtask conditioning aligns value resets with predicted subtask changes.}
    Both critics evaluate recorded dataset action chunks.
    Top: predicted values along a held-out
    \textbf{shirt-hang} trajectory. Bottom: enlarged views of the five shaded regions;
    colored dashed lines mark each critic's predicted switch nearest the panel center.
    \emph{Fall span} counts frames between the first $5\%$ and $95\%$ of the local
    peak-to-trough drop, using the maximum value in the $20$ frames before that
    critic's switch and the minimum from the switch through $20$ frames after it.
    A drop crossing both thresholds in one frame has span $0$. Panel titles report
    \methodname{} versus no conditioning.}}
    \label{fig:qualitative}
\end{figure}

Figure~\ref{fig:qualitative} evaluates both critics on the same held-out
\textbf{shirt-hang} trajectory, using the same checkpoints
as in our evaluations. We also decode
the ablation's subtask prediction for this diagnostic, without supplying it to the value head. Although
the overall value curves are similar, their behavior near subtask changes differs.
For \methodname{}, the largest one-frame drop coincides with the predicted subtask
switch in all five displayed transitions, and both fall thresholds are crossed in
that frame. Without conditioning, the drops span $4$--$24$ frames and can begin before
or finish after the switch. These results suggest that, without explicit subtask
conditioning, the model learns two highly accurate but largely independent
predictions. The lags between subtask switches and value resets indicate that
these predictions lack the desired coupling. In \methodname{}, explicit conditioning
ties the value reset to a change in the model's inferred subtask.

\subsubsection{Value Predictions across Learning Objectives}
\label{app:value_viz}
The objectives in Section~\ref{sec:baselines} differ both in their backup and in the
horizon over which reward must propagate. Figure~\ref{fig:value-overestimation}(b)
shows persistent underestimation and a broad distribution of $Q-\mathrm{MC}$
differences for task-level TD-BoN during RoboCOIN pretraining. One plausible source
of this dispersion is variation in task horizon: with sparse terminal rewards,
TD estimates can collapse toward zero on very long tasks while remaining useful
on shorter tasks.

Figure~\ref{fig:downstream-value-generalization} compares snapshot-and-value
strips on \textbf{shirt-hang} and \textbf{grocery-packing} trajectories to
qualitatively assess the generalization of the learned value functions.

\begin{figure}[p]
    \centering
    \input{figures/downstream_generalization/figure}
\end{figure}

Figure~\ref{fig:pretraining-value-generalization} compares \methodname{} with
the no-pretraining and task-specific ResNet-50 baselines from Section~\ref{sec:pretraining} on a
held-out \textbf{shirt-hang} trajectory.

\begin{figure}[t]
    \centering
    \input{figures/downstream_generalization/pretraining_figure}
\end{figure}

\subsection{Hyperparameters}
\label{sec:hparams}

\textbf{Shared critic settings.} Except for the ResNet-50 ablation, all critics use
a PaliGemma backbone with $224\times224$ images from three cameras and
\texttt{float32} precision. Critics use AdamW with weight decay $10^{-6}$,
$\beta_1=0.9$, $\beta_2=0.95$, $\epsilon=10^{-8}$, and gradient-norm clipping at $1$.
TD critics use a target-network EMA rate of $\tau=0.005$.
We use the action representation and quantile normalization described in
Appendix~\ref{app:action_space}, with action chunks of length $50$ for RoboCOIN
pretraining and $60$ for downstream training. RoboCOIN critic pretraining uses
batch size $256$; all downstream critic training uses batch size $128$.

\textbf{Base policies.} For each downstream task, we finetune $\pi_{0.5}$ from its
released base checkpoint. The same fixed task-specific policy supplies candidates
for \methodname{} and all critic ablations, and is evaluated directly as the BC
baseline. Policies use Adam and the same optimizer
coefficients and clipping threshold as the critics. Their cosine learning-rate
schedules have peak $5\times10^{-5}$, $1000$ warmup steps, and endpoint
$5\times10^{-6}$. Table~\ref{tab:hp-policy} distinguishes the number of training
steps used for the selected policy from the full LR schedule length. The grocery-packing and
LEGO-disassembly policies condition on manually advanced subtask prompts; this is a property
of the base policy, separate from the critic's own subtask prediction.

\begin{table}[ht]
\centering
\caption{\footnotesize{\textbf{Downstream base-policy hyperparameters.}
Training steps refer to the policy checkpoint used in evaluation; LR decay length
refers to the full configured cosine schedule.}}
\label{tab:hp-policy}
\begin{tabular}{lccc}
\toprule
\textbf{Task} & \textbf{Training steps} & \textbf{LR decay steps} & \textbf{Batch size} \\
\midrule
\textbf{shirt-hang}  & 60k & 200k & 128 \\
\textbf{lid-sealing} & 40k & 60k  & 256 \\
\textbf{grocery-packing}     & 70k & 70k  & 256 \\
\textbf{LEGO-disassembly}        & 20k & 20k  & 256 \\
\bottomrule
\end{tabular}
\end{table}

The separate $\pi_{0.5}$ policy used to generate candidate actions for RoboCOIN
critic pretraining is trained for $230$k steps with batch size $256$. Its learning
rate warms up for $1000$ steps to $10^{-5}$ and then remains constant.

\textbf{Critic pretraining.} All PaliGemma critics in Table~\ref{tab:hp-vf-base}
are pretrained on the same $131$ RoboCOIN task datasets for $230$k steps, using a
cosine learning-rate schedule with peak $10^{-5}$, $1000$ warmup steps, and endpoint
$10^{-6}$. The auxiliary next-token prediction (NTP) loss has weight $0.1$ whenever
subtask prediction is enabled, and $0$ otherwise. TD-BoN uses $8$ policy candidates
for its pretraining backup.

\begin{table}[ht]
\centering
\caption{\footnotesize{\textbf{Critic pretraining variants.} All rows use $230$k
pretraining steps and the shared schedule above. Conditioning refers to the
critic's value prediction, independently of the base-policy prompt.}}
\label{tab:hp-vf-base}
\resizebox{\linewidth}{!}{%
\begin{tabular}{llccc}
\toprule
\textbf{Critic} & \textbf{Return scope} & \textbf{$\gamma$} & \textbf{NTP weight} & \textbf{Subtask conditioning} \\
\midrule
\methodname{} (TD-BoN)       & Subtask & 0.999  & 0.1 & Yes \\
SARSA                        & Subtask & 0.999  & 0.1 & Yes \\
MC                           & Task    & 0.9995 & 0   & No \\
TD-BoN                       & Task    & 0.9995 & 0   & No \\
NTP, no conditioning         & Subtask & 0.999  & 0.1 & No \\
Neither NTP nor conditioning & Subtask & 0.999  & 0   & No \\
\bottomrule
\end{tabular}
}
\end{table}

\textbf{Downstream critic finetuning.} Each pretrained critic starts from its
$230$k checkpoint. Finetuning uses a cosine LR from $5\times10^{-6}$ to
$5\times10^{-7}$ over $20$k steps, without warmup. Table~\ref{tab:hp-vf-ft} lists
the selected checkpoints: $250$k corresponds to $20$k finetuning steps, while
$240$k corresponds to $10$k finetuning steps within the same $20$k LR schedule.
For the \textbf{shirt-hang} variant with NTP but no subtask conditioning, we select
the $240$k checkpoint because it performed better than the $250$k checkpoint.
This differs from the evaluation protocol in the remaining experiments, where
we do not perform checkpoint selection. The resulting success rate can therefore
be viewed as a favorable upper bound for this ablation; the conclusion that
explicit subtask conditioning improves performance still holds. This value
function is used for policy-steering comparisons only in
Section~\ref{sec:subtask-ablation}.
For \methodname{} on \textbf{LEGO-disassembly}, we use the $240$k checkpoint
because this task has a much smaller dataset.

\begin{table}[ht]
\centering
\caption{\footnotesize{\textbf{Downstream critic checkpoints.} All rows initialize
from the corresponding $230$k pretrained critic and use a $20$k LR decay schedule.}}
\label{tab:hp-vf-ft}
\resizebox{\linewidth}{!}{%
\begin{tabular}{llcc}
\toprule
\textbf{Critic} & \textbf{Downstream task} & \textbf{Finetuning steps} & \textbf{Checkpoint step} \\
\midrule
\methodname{}                & \textbf{shirt-hang}, \textbf{lid-sealing}, \textbf{grocery-packing} & 20k & 250k \\
\methodname{}                & \textbf{LEGO-disassembly}                                            & 10k & 240k \\
Task-level MC                & \textbf{shirt-hang}, \textbf{grocery-packing}                    & 20k & 250k \\
Task-level TD-BoN            & \textbf{shirt-hang}, \textbf{grocery-packing}                    & 20k & 250k \\
Subtask-level SARSA          & \textbf{shirt-hang}, \textbf{grocery-packing}                    & 20k & 250k \\
NTP, no conditioning         & \textbf{shirt-hang}                                     & 10k & 240k \\
Neither NTP nor conditioning & \textbf{shirt-hang}                                     & 20k & 250k \\
\bottomrule
\end{tabular}%
}
\end{table}

\textbf{Critics without robot pretraining.} The two task-specific
\textbf{shirt-hang} ablations are trained directly for $20$k steps and evaluated
at checkpoint $20$k. \methodname{} without robot pretraining initializes from
PaliGemma and uses a cosine LR from $5\times10^{-6}$ to $5\times10^{-7}$ over
$20$k steps with $1000$ warmup steps. The ResNet-50 + MLP critic initializes its
image encoder from ImageNet weights and uses a cosine LR from $10^{-5}$ to
$10^{-6}$ over $20$k steps with $1000$ warmup steps. Both use the subtask-level
TD-BoN objective with $\gamma=0.999$, $8$ backup candidates, and subtask-prediction
loss weight $0.1$; the ResNet critic predicts the subtask as a categorical label.

\subsection{RoboCOIN Pretraining Dataset}
\label{app:robocoin-data}
\begin{figure}[H]
    \centering
    \includegraphics[width=0.58\linewidth]{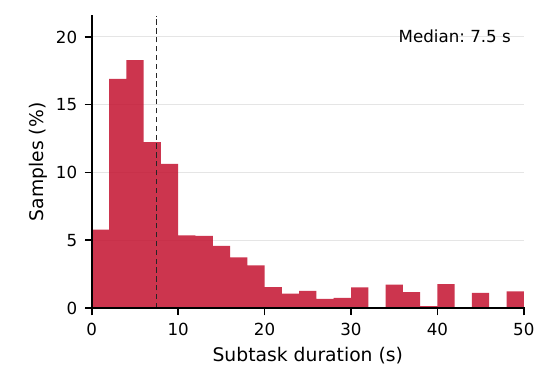}
    \vspace{-0.2cm}
    \caption{\footnotesize{\textbf{Histogram of RoboCOIN subtask durations in seconds.}}}
    \label{fig:subtask-distribution}
\end{figure}

We construct our diverse pretraining dataset from the bimanual-embodiment subset of
RoboCOIN~\citep{wu2026robocoinopensourcedbimanualrobotic}, including humanoid robot
data. The combined training and validation splits contain $131$ task datasets,
$40{,}054$ episodes, and $31{,}337{,}006$ steps across three embodiments: Agilex Cobot
Magic, Agilex Split ALOHA, and Galaxea R1 Lite. This is the filtered set available
when we constructed the dataset; subsequent RoboCOIN releases have added more data.
Table~\ref{tab:robocoin-tasks} lists the task datasets included in this snapshot.
Figure~\ref{fig:subtask-distribution} shows the distribution of subtask durations in
RoboCOIN, in seconds. To match the temporal scale of pretraining, we define
subtasks in the target tasks with average durations of $\approx5$--$25$ seconds.

\clearpage
\begingroup
\small
\setlength{\tabcolsep}{5pt}
\renewcommand{\arraystretch}{1.1}
\begin{longtable}{@{}*{3}{>{\raggedright\arraybackslash}p{0.315\linewidth}}@{}}
\caption{\footnotesize{\textbf{The $131$ RoboCOIN task datasets used for pretraining.}
Task identifiers are grouped by embodiment, with the embodiment prefix omitted
and underscores rendered as hyphens. Released spellings and variant suffixes are
retained.}}\phantomsection\label{tab:robocoin-tasks}\\
\toprule
\endfirsthead
\multicolumn{3}{l}{\tablename~\thetable\ (continued)}\\
\toprule
\endhead
\bottomrule
\endfoot
\multicolumn{3}{@{}l}{\textbf{Agilex Cobot Magic (63 tasks)}}\\*
\midrule
\textbf{box-storage-chopsticks} & \textbf{cap-the-pen-a} & \textbf{catch-the-ball} \\
\textbf{classification-of-fruits-and-vegetables} & \textbf{classification-of-fruits-and-vegetables-a} & \textbf{classification-of-tableware} \\
\textbf{clean-blackboard} & \textbf{clean-up-the-tableware} & \textbf{clear-the-desktop} \\
\textbf{close-book} & \textbf{close-button} & \textbf{cube-reset} \\
\textbf{cut-banana} & \textbf{desktop-organization} & \textbf{drawer-storage-mineral-water} \\
\textbf{fold-clothes} & \textbf{fold-the-towel} & \textbf{fold-towel-a} \\
\textbf{food-packaging} & \textbf{make-fruit-salad} & \textbf{make-hamburger} \\
\textbf{mobile-cube} & \textbf{mobile-cube-blackboard} & \textbf{move-beverage} \\
\textbf{move-plate} & \textbf{move-the-ball} & \textbf{move-the-ball-and-the-cube-block} \\
\textbf{move-the-ball-interference} & \textbf{move-the-bread} & \textbf{move-the-cup} \\
\textbf{move-the-plate} & \textbf{move-the-small-ball} & \textbf{movethe-position-of-the-bluetooth} \\
\textbf{open-the-shoebox} & \textbf{place-square-pyramid} & \textbf{place-the-cube-block} \\
\textbf{place-the-test-tube} & \textbf{plate-storage-apple} & \textbf{plate-storage-bread} \\
\textbf{plate-storaje-baozi} & \textbf{pot-storage-steamer} & \textbf{pour-drink} \\
\textbf{pour-water-a} & \textbf{pour-water-bottle} & \textbf{prepare-breakfast} \\
\textbf{pull-zipper} & \textbf{pushing-magnet} & \textbf{put-in-the-pear} \\
\textbf{put-the-building-block-on-the-table} & \textbf{steamer-storage-dumpling} & \textbf{storage-plate} \\
\textbf{take-out-a-pen-from-the-pen-holder} & \textbf{take-out-the-bread} & \textbf{take-the-shoes-off-the-shelf} \\
\textbf{the-box-stores-table-tennis-balls} & \textbf{the-plate-holds-the-fruit} & \textbf{the-plate-holds-the-vegetables} \\
\textbf{turn-off-the-desk-lamp} & \textbf{turn-on-the-bulb} & \textbf{turn-on-the-desk-lamp} \\
\textbf{twist-bottle-cap} & \textbf{vase-storage-flower} & \textbf{water-bottle-storage} \\
\addlinespace[5pt]
\midrule
\multicolumn{3}{@{}l}{\textbf{Agilex Split ALOHA (16 tasks)}}\\*
\midrule
\textbf{basket-storage-banana} & \textbf{basket-storage-bread} & \textbf{basket-storage-egg-yolk-pastry} \\
\textbf{basket-storage-long-bread} & \textbf{basket-storage-orange} & \textbf{basket-storage-peach} \\
\textbf{fold-the-pants} & \textbf{plate-storage} & \textbf{pour-rice} \\
\textbf{pour-tea} & \textbf{scoop-coffee-beans} & \textbf{stack-baskets} \\
\textbf{stir-coffee} & \textbf{wipe-table} & \textbf{wipe-the-table} \\
\textbf{zip-up-the-document-bag} &  &  \\
\addlinespace[5pt]
\midrule
\multicolumn{3}{@{}l}{\textbf{Galaxea R1 Lite (52 tasks)}}\\*
\midrule
\textbf{boil-water-in-a-kettle} & \textbf{catch-the-water} & \textbf{clean-the-floor} \\
\textbf{clean-the-sink} & \textbf{clean-toilet} & \textbf{connect-the-router-cable} \\
\textbf{cook-a-meal} & \textbf{cover-the-pot-lid} & \textbf{dispose-of-leftover-food} \\
\textbf{drawer-storage-hair-dryer} & \textbf{fold-clothes} & \textbf{garbage-disposal} \\
\textbf{hang-clothes} & \textbf{make-a-landline-call} & \textbf{make-breakfast} \\
\textbf{make-tea} & \textbf{make-the-bed} & \textbf{open-and-close-curtains} \\
\textbf{open-and-close-microwave-oven} & \textbf{open-and-close-nightstand-door} & \textbf{open-and-close-nightstand-drawer} \\
\textbf{open-and-close-the-freezer-door} & \textbf{open-the-food-pan} & \textbf{opening-and-closing-aalcony-sliding-doors} \\
\textbf{pick-up-and-store-items} & \textbf{place-the-dress-shirt-on-the-hanger} & \textbf{plug-the-socket} \\
\textbf{pour-water} & \textbf{put-on-a-garbage-bag} & \textbf{put-slippers-into-floor-standing-shoe-cabinet} \\
\textbf{put-the-pillow-on-the-bed} & \textbf{put-the-shoes-into-the-shoe-box} & \textbf{put-the-tableware-into-the-cupboard} \\
\textbf{sliding-chair} & \textbf{storage-of-toiletries} & \textbf{switch-labels} \\
\textbf{switch-on-and-off-the-central-air-conditioning} & \textbf{tableware-arrangement} & \textbf{tableware-cleaning} \\
\textbf{take-and-place-the-portable-power-bank} & \textbf{take-and-put-away-garden-stuff} & \textbf{take-and-put-away-garden-stuff-a} \\
\textbf{take-and-put-away-items} & \textbf{take-and-put-the-bowl} & \textbf{take-clothes-out-of-the-washing-machine} \\
\textbf{take-or-store-plates} & \textbf{tea-service-table-setting} & \textbf{throw-out-the-trash} \\
\textbf{tidy-up-toiletries} & \textbf{wash-the-tableware} & \textbf{washing-board} \\
\textbf{wipe-the-table} &  &  \\
\addlinespace[5pt]
\end{longtable}
\endgroup

\end{document}